# Probabilistic Prognostic Estimates of Survival in Metastatic Cancer Patients (PPES-Met) Utilizing Free-Text Clinical Narratives

Imon Banerjee, Ph.D.[1], Michael Francis Gensheimer, MD[2], Douglas J. Wood, MS[1], Solomon Henry, MS[1], Daniel Chang, MD[2], Daniel L. Rubin, MD[1]
[1]Department of Biomedical Data Science, Stanford University, Stanford, CA,
[2]Department of Radiation Oncology, Stanford University, Stanford, CA

**Introduction**

Scientific review(1) and clinical studies(2) have indicated that physicians are overly-optimistic during survival prediction of patients with terminal cancer. This leads to overutilization of aggressive medical interventions and protracted radiation treatment, which increase side effects and health care bills, while other patients who may benefit for continued therapy are likely under-treated. The electronic medical records systems serving major medical centers store a wealth of potentially valuable clinical notes acquired over time ("longitudinal data") that contains rich information on performance status, imaging findings, and tolerance to systemic therapy. These data contain key covariates for informing diagnostic decisions, and may also serve as critical resources for survival estimation. However, the massive explosion of the medical data outstrips the manual ability to comprehend the entire source of information. On the contrary, a machine learning model can be trained on the large amount of heterogenous data which allows seamless integration between multi-source information, and, as a result, it may outperform physicians in estimating life expectancy. Using the large Stanford Cancer Institute Research Database (SCIRDB), we created a dynamic sequence-dependent deep learning model - *Probabilistic Prognostic Estimates of Survival in Metastatic Cancer Patients (PPES-Met)*. The *PPES-Met* model takes as input a sequence of longitudinal free-text clinical visit narratives ordered according to the date of visits, and computes as output a probability of short-term life expectancy (> 3 months) for each visit. The complexity of the model mainly lies in extracting relevant information from the heterogeneous types of free-text clinical notes along with modeling temporal irregularity of the visits.

**Methods**

*Patient population:* With the approval of Institutional Review Board (IRB), we created a database that includes adult cancer patients (13,523) seen at the Stanford Cancer Center from 2008-2017 and diagnosed with distant metastases. This database contains various types of free-text patient visit notes (e.g. radiology reports, oncologist notes, discharge summary) from date of metastatic cancer diagnosis to death. A separate "*Palliative radiation dataset*" was created using patients (899) enrolled from 2015-2016 in a prospective survey study conducted in our institution's Radiation Oncology department. The overall group of patients were seen for 471,005 daily encounters/visits, including outpatient and inpatient contact. For these visits, median follow-up was 12.7 months. Median overall survival was 22.4 months. Patients were hospitalized for 115,716 (24.6%) visits. There were 1,403,544 provider notes. The training set contains 10,239 patients with 380,080 visits, validation set of 1,785 patients and test set of 1,818 patients (15%) with 90,925 visits.

*Proposed System - PPES-Met:* The model is composed of two core processing blocks:

*i. Semantic Word Embedding (SWE):* We adopted a completely unsupervised hybrid method – an updated version of *Intelligent Word Embedding (IWE)* method(3) that combines semantic-dictionary mapping, neural embedding, and context-based windowing technique for creating dense vector representation of free-text clinical narratives. The method leverages the benefits of unsupervised learning along with expert-knowledge to tackle the major challenges of information extraction of informative information from clinical texts, while accounting for ambiguity of free text narrative style, lexical variations, arbitrary ordering of words, and frequent appearance of abbreviations and acronyms.

*ii. Stacked RNN model:* On top of the word embeddings obtained from patient visit notes, we designed a many-to-many RNN model using two-layer one directional stacked stateful Long short-term memory (LSTM) units for learning survival across the sequence of clinical narratives. The model takes as input a series of vectorized patient visit notes ordered according to the timestamp of visits, and it predicts probability of survival at each patient visit. In the stacked RNN layers, the first layer's one directional LSTM block receives the input $x^t$ and previous hidden state $h_{t-1}$, and pass the current state $h_t$ to the successive LSTM block and to the corresponding block in the upper layer. LSTM blocks also maintains an H-dimensional cell state $C_{t-1} \in R^H$ and the second layer units have modeled to maintain the recurrent connections in multiple dimensions. The final output of double layer stacked-RNN is modeled as: $\hat{y}_t^{(i)} = f_h(h_t^i, \hat{y}_{t-1}^{(i)})$ where $h_t^i$ is the hidden state of the first level and $\hat{y}_t^{(i)}$ is the predicted survival at time $t - 1$. The time distributed weighted cross entropy loss is used as optimization function during model training. By using stateful LSTM, the model includes long-term dependencies that exist in the longitudinal data, which is generally very informative for the prediction task.

## Results

*Quantitative evaluation:* We used 3-month survival record defined at each time point as categorical class labels for probabilistic prediction. The overall prognosis estimation accuracy on the test dataset was validated as Precision-Recall curve and AUC score was 0.97. The high AUC scores show that the model is predicting survival accurately with high precision, as well as returning most of positive results (high recall). To check the calibration with the ground truth, we also measure the Brier score and the low brier-score (0.069) shows the prediction was highly calibrated with the ground truth.

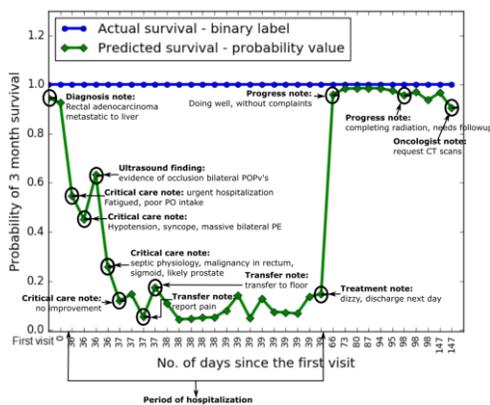

**Figure 1** Intelligible longitudinal survival curve of a patient

*Qualitative evaluation:* To provide a formalized mechanism to reason about computerized model's predictions at a specific timepoint, we implemented an interactive graphical interface that generates a longitudinal probabilistic summary (Figure 1) for each patient. Clicking on a time point, the system will retrieve not only the visit type, but it will also exploit the controlled-terms and will extract the core findings of the visit by highlighting the context of the controlled-terms (see Figure 1). This intuitive illustration may help the clinician to reason on the PPE-Met prediction and perform a qualitative error-analysis.

## Discussion

The objective of our work is to improve physicians' knowledge of their patients' prognosis to help tailor treatment strategy, improve quality of life, as well as reduce costs. Early stage results of an ongoing prospective study conducted by Stanford Radiation Oncology department which includes 899 patients enrolled in the palliative radiation study found that the physicians are not able to accurately estimate life expectancy. We tested our *PPES-Met* model on a combination of a general group of metastatic patients and data from a palliative radiation study, and the probabilistic prediction accuracy was 0.97 AUC-PR-curve. Initial evaluation suggests that the PPES-Met prediction model produces good accuracy. This is probably due to the *PPES-Met* model's capability of integrating a large amount of patient-specific facts while preserving long-term dependencies in the data, which is not a trivial task for human experts. The high accuracy and the ability of our visualization the *PPES-Met* model output to show the critical data driving its predictions suggests that the model might ultimately be clinically useful in the future as a decision support tool to personalize metastatic cancer treatment and to aid physicians. The core limitations of the current work are: (i) many patients in the training dataset lost follow-up which may create an inaccurate assumption about survival during the model training. However, we consulted the central cancer registry to validate the survival; (ii) the time points are not equally spaced, and a single day may contribute multiple data points which may affect the temporal dependency and introduce fluctuation in the survival curve. In future, an NLP technique can be applied to combine the visit data from the same day for date-based survival analysis.